\documentclass[english]{bvm2020}

\def\articlenumber{0000}

\usepackage{amsmath} % Required for \varPsi below
\usepackage{tikz}
\usepackage{pgfplots}

\date{}
\title{Epoch-wise label attacks for robustness against label noise}

\subtitle{Chest X-ray tuberculosis classification with corrupted labels}

\author{Sebastian~G\"undel$^1$, Andreas~Maier$^1$}
\authorrunning{G\"undel et al.}%Guendel et al.}
\institute{$^1$Pattern Recognition Lab, FAU Erlangen-N\"urnberg\\}%
\email{sebastian.guendel@fau.de}%sebastian.guendel@fau.de}

\begin{document}

%==============================================================================
% wählen Sie mit dem Befehl \selectlanguage die Sprache aus, in der Ihr 
% Proceeding verfasst ist
%
%\selectlanguage{german}
\selectlanguage{english}

\maketitle

\begin{abstract}
The current accessibility to large medical datasets for training convolutional neural networks is tremendously high. The associated dataset labels are always considered to be the real ``ground truth". However, the labeling procedures often seem to be inaccurate and many wrong labels are integrated. This may have fatal consequences on the performance of both training and evaluation. In this paper, we show the impact of label noise in the training set on a specific medical problem based on chest X-ray images. With a simple one-class problem, the classification of tuberculosis, we measure the performance on a clean evaluation set when training with label-corrupted data. We develop a method to compete with incorrectly labeled data during training by randomly attacking labels on individual epochs. The network tends to be robust when flipping correct labels for a single epoch and initiates a good step to the optimal minimum on the error surface when flipping noisy labels. On a baseline with an AUC (Area under Curve) score of 0.924, the performance drops to 0.809 when 30\% of our training data is misclassified. With our approach the baseline performance could almost be maintained, the performance raised to 0.918.

\end{abstract}

\section{Introduction}
Current research highlights the vast number of datasets where corresponding labels are partly incorrect. In the medical field this can be caused by many different reasons, e.g., errors in the labeling procedure when retrieving from the clinical reports. In addition, radiologists may misinterpret clinical images which lead to incorrect ground truth \cite{radiology_errors}.\par
Different strategies can be applied to handle datasets with noisy labels: Additional radiologists re-annotate the dataset labels to check the variability between the original labels and the radiologists \cite{Rajpurkar2018DeepLF}. However, most datasets contain an extensive number of images and a small fraction can only be processed. Dealing with label noise in the training set, robust loss functions are generated \cite{Zhang2018GeneralizedCE}. \\

In this paper we implement a robust method to deal with this label noise in the training set on a binary class problem. By randomly attacking labels in each epoch, a fraction of noisy labels may be switched such that the network is trained with correct labels. We use tuberculosis classification based on CXR images and artificially insert label noise to analyse the effects of different noise ratios. 

\section{Materials and methods}
\subsection{Datasets}

For pulmonary tuberculosis classification based on chest X-ray images, there are two public datasets available. The first, the Montgomery dataset contains 138 frontal images, 58 with and 80 without tuberculosis. The second dataset is derived from the Shenzhen hospital including 662 X-rays. Half of the images include tuberculosis, the other half has no evidence. The entire collection contains 800 CXR images \cite{datasets}. \\
For training purposes, we downsample all images to 256$\times$256. We split the data collection into 70\% for training, 10\% for validation, and 20\% for testing. For the experiments, we treat the corresponding dataset labels as clean without any ratio of noise.

\begin{figure}[t]
\centering
\includegraphics[width=10cm]{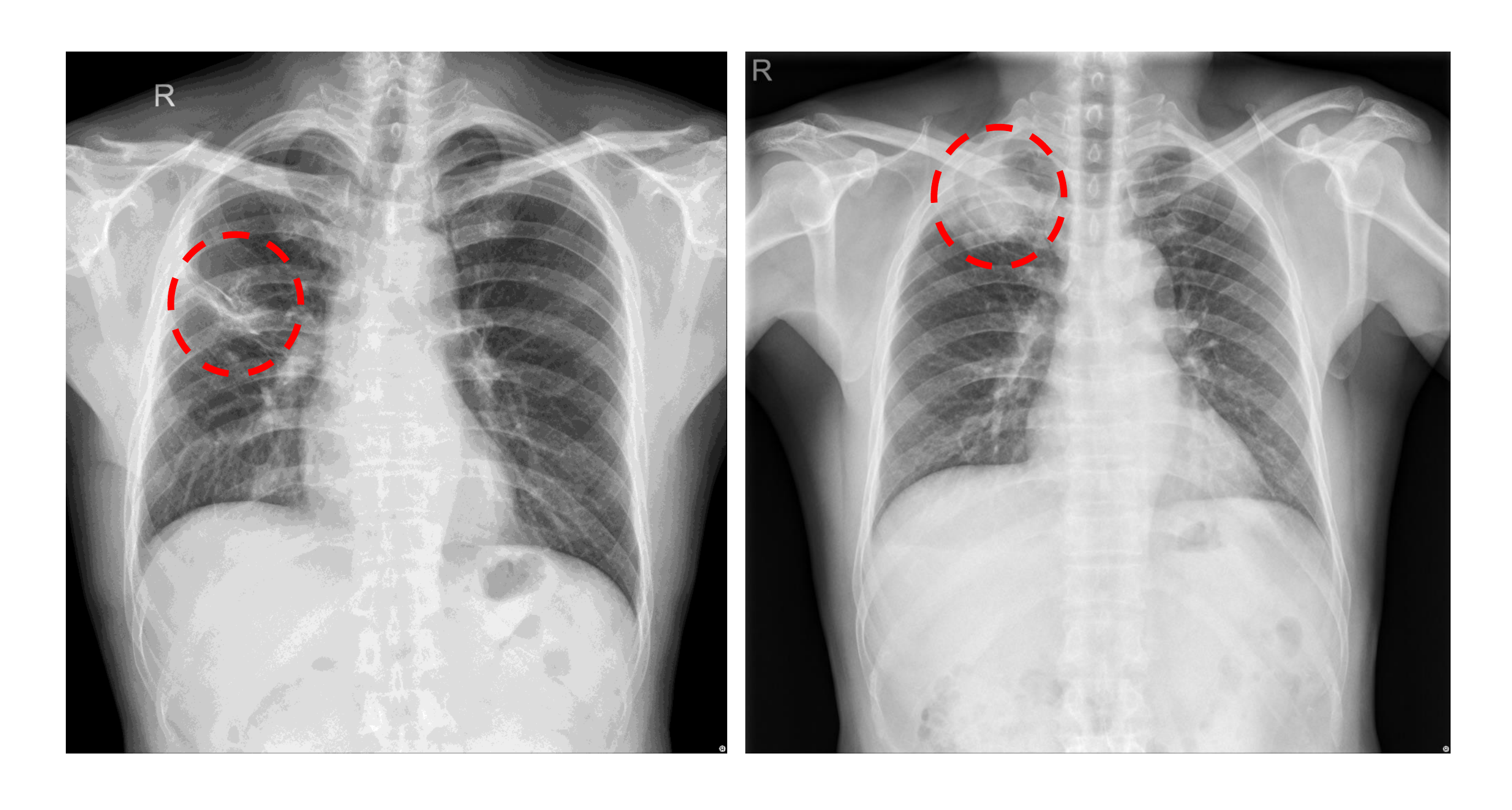}
\vspace{-0.3cm}
\caption{Two example chest X-ray images with tuberculosis; Tuberculosis in the right middle lung (left) and tuberculosis in the right upper lung (right). More difficult cases may result in a wrong classification.}
\end{figure}

\subsection{Network and Training Setup}

As convolutional neural network, we use a densly-connected model (DenseNet) with 121 layers \cite{Huang2016DenselyCC}. We load the ImageNet pretrained weights before starting the training. The input image is accordingly normalized and provided in the three input channels. The output layer of the network is reduced such that one sigmoidal unit is returned. During training we use the Adam optimizer \cite{adam} ($\beta_1$ = 0.9, $\beta_2$ = 0.999) and a learning rate of $10^{-4}$. We stop the training and jump back to the best epoch if the validation accuracy does not improve after a patience of 8 epochs. We apply the binary cross-entropy function to predict the loss. Each batch is filled with 16 examples. The performance on the test set is evaluated with the area under ROC curve (AUC).

\subsection{Label Noise}

Many medical datasets include incorrectly labeled data. As we assume to have a clean dataset, we artificially inject a portion of label noise to the data and measure the performance to see the effects on the performance. For all experiments, label noise is only applied to the training set. The labels of the validation set are kept such that we can retrieve our best model. An important factor to evaluate a model is to measure the performance on clean labels. Often, this clean test set can not be guaranteed as the whole dataset is corrupted. Since we artificially integrate label noise on the training data only, our model can be correctly evaluated and the returned performance scores can be considered as valid.\\

Assuming we have the clean labels $l_i$ for all examples $i$, we define two flip probabilities $p_p=p(\hat{l}_i=0|l_i=1)$ and $p_n=p(\hat{l}_i=1|l_i=0)$. The two parameters describe how many labels are incorrect before we start the training process. In our experiments we simplify the problem such that the same ratio of positive and negative cases are flipped: $p_1(e=1)=p_p=p_n$. First, we train our baseline model on the clean, original labels. We reach an AUC of 0.924. Inducing higher ratios of label noise the performance drops as can be seen in Table \ref{tab:noise_performance}. Even if a high amount of labels are incorrect (e.g. $p_1(e=1)=0.5$), the model can classify many examples correctly.

\begin{table}[h]
\begin{center}
\begin{tabular}{l | c c c c c c} 

$p_1(e=1)$ \hspace{1cm}& \hspace{5mm}0.0 & \hspace{5mm}0.1 & \hspace{5mm}0.2 & \hspace{5mm}0.3 & \hspace{5mm}0.4 & \hspace{5mm}0.5 \\
\hline
AUC & \hspace{5mm}0.924 & \hspace{5mm}0.894 & \hspace{5mm}0.835 & \hspace{5mm}0.809 & \hspace{5mm}0.791 & \hspace{5mm}0.775\\

\end{tabular}
\\
\end{center}
\caption{AUC score when the labels were flipped with probability $p_1$ before training.}
\label{tab:noise_performance}
%\vspace{-0.15in}
\end{table}

\subsection{Individual Label Attacks}
The model is widely robust to a certain amount of label noise. We hypothesize that individual, epoch-wise label attacks make the model even more stable in terms of the classification performance. Therefore, we define a new probability $p_2(e=1)$, which changes the label for a single epoch only. In this case the probability is tremendously smaller that a label is incorrect for the entire training process. We measure the performance based on different epoch-wise noise ratios $p_2(e=1)$ when we have no prior label noise ($p_1(e=1)=0$).

\begin{table}[h]
\begin{center}
\begin{tabular}{l | c c c c c c}

$p_2(e=1)$ \hspace{1cm}& \hspace{5mm}0.1 & \hspace{5mm}0.2 & \hspace{5mm}0.3 & \hspace{5mm}0.4& \hspace{5mm}0.5& \hspace{5mm}0.6\\
\hline
AUC  & \hspace{5mm}0.901 & \hspace{5mm}0.917 & \hspace{5mm}0.888 & \hspace{5mm}0.905 & \hspace{5mm}0.639 & \hspace{5mm}0.475 \\

\end{tabular}
\\
\end{center}
\caption{AUC score when the labels were flipped in each epoch with probability $p_2$.}
\label{tab:noise_performance2}
%\vspace{-0.15in}
\end{table}

In Table \ref{tab:noise_performance2}, we see that training with individual, epoch-wise attacks is significantly more robust than constant noise over the examples. Even if we flip in each epoch with a probability $p_2(e=1)=0.4$, the performance can nearly reach the baseline performance. However, for experiments with $p_2(e=1)\ge0.5$, the performance significantly drops. 

\begin{figure}[h]
\centering
\includegraphics[width=9cm]{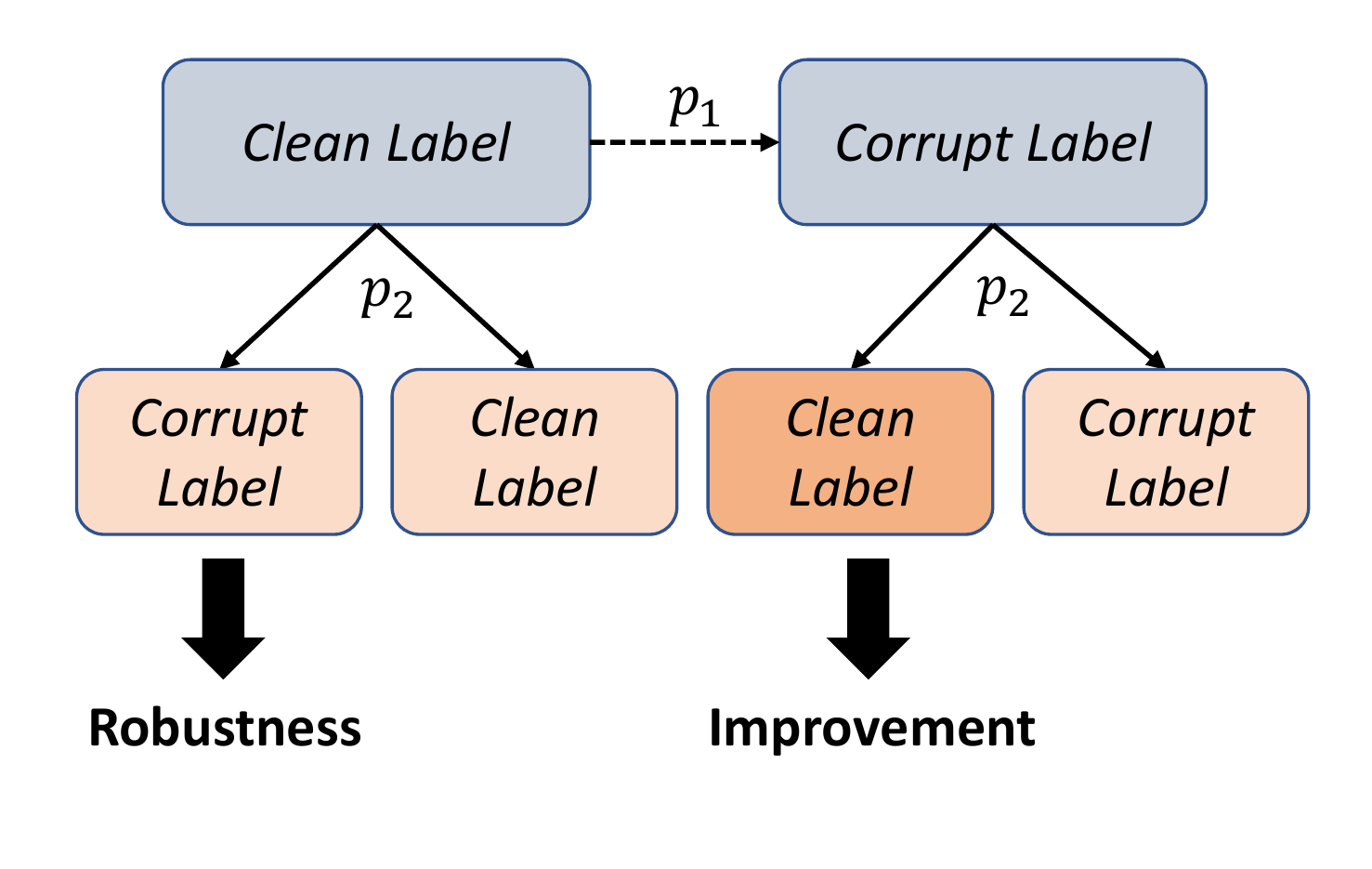}
\vspace{-0.5cm}
\caption{Label Attack Strategy: The dataset includes prior noise with a certain noise rate (blue blocks). Flipping a label in one epoch ends up in one of the four scenarios: A corrupted/clean label derived from a clean label or a corrupted/clean label derived from a corrupted label (red blocks).}
\label{fig:label_attack}
\end{figure}

\textbf{Individual Label Attacks on Prior Label Noise: } The main goal of this paper is to show that the individual and epoch-wise label attacks help to improve the classification performance when prior label noise is integrated in the training set. Figure \ref{fig:label_attack} visualizes the four scenarios which are possible for each label based on prior noise with $p_1$ and the epoch-wise label flips with $p_2$.

\textbf{Probability \textit{p\textsubscript{2}} derivation:}
In most datasets, the label noise ratio is unknown. For the determination of the label flip probability $p_2$, we can derive the value without knowledge of the label noise ratio. For the $p_2$ determination we define a sample flip minimum and maximum. This can be derived from the binomial distribution

\begin{equation}
    B(k|p,n)=\frac{n!}{k! \cdot (n-k)!}\cdot p^k(1-p)^{n-k},
\label{eq:binomial}
\end{equation}

where $k$ is the number of flips, $p=p_2$ the flip probability, and $n$ the number of epochs. We use our previous experiments to see how many epochs are trained. An average training duration of 18 epochs is predicted ($n=18$). For the prediction of the optimal $p_2$, we define 2 constraints:

\begin{itemize}
    \item An example should be flipped at least once in the training ($B(k=0)\approx0)$
    \item An example should be flipped less than a half in the training ($B(k\ge\frac{n}{2})\approx0$)
\end{itemize}

For simplification for the prediction $p$, we say that 

\begin{equation}
    B(k_1=0)=B(k_2=\frac{n}{2}).
\label{eq:binomial2}
\end{equation}

This condition is fulfilled if the mean $\mu = p*n = \frac{k_1+k_2}{2}$. Thus, we can predict the epoch-wise flip probability

\begin{equation}
    p_2=\frac{k_1+k_2}{2n} = \frac{9}{36} = 0.25.
\label{eq:p2}
\end{equation}

According to Equation \ref{eq:binomial}, we get a probability $B(k_1=0)=6*10^{-3}$ that an example is never flipped in the training and a probability $B(k_3\ge\frac{n}{2})=\sum_{k=\frac{n}{2}}^n {n \choose k} p^k(1-p)^{n-k}=0.0193$ that an example is flipped in a half or more epochs. We hypothesize that this fraction of examples may not contribute to the training improvement under the condition that we have a noisy label for $B(k_1=0)$ and a clean label for $B(k_3\ge\frac{n}{2})$. Thus, the real probability that an example may not contribute is a multiplication with $p_1$ or $(1-p_1)$, respectively, which results in a significantly smaller value.

\section{Results}

\begin{table}[h]
\begin{center}
\begin{tabular}{l | c c c c c c} 

$p_1(e=1)$ \hspace{1cm}& \hspace{5mm}0.1 & \hspace{5mm}0.2 & \hspace{5mm}0.3 & \hspace{5mm}0.4 & \hspace{5mm}0.5 \\
\hline
AUC & \hspace{5mm}0.891 & \hspace{5mm}0.883 & \hspace{5mm}0.918 & \hspace{5mm}0.846 & \hspace{5mm}0.846\\
AUC gain & \hspace{5mm}-0.003 & \hspace{5mm}+0.048& \hspace{5mm}+0.109 & \hspace{5mm}+0.055 & \hspace{5mm}+0.071\\

\end{tabular}
\\
\end{center}
\caption{AUC scores when the dataset labels were flipped with a certain probability $p_1$ before training and attacked with an epoch wise flip probability $p_2=0.25$.}
\label{tab:noise_performance3}
%\vspace{-0.15in}
\end{table}

We evaluated our method with a constant $p_2$ defined in Equation \ref{eq:p2} under varying noise probabilities $p_1$. Table \ref{tab:noise_performance3} shows the performance on the evaluation. We apply the noise on the same labels as is the experiments in Table \ref{tab:noise_performance}. The best performance gain could be achieved with $p_1=0.3$, from an AUC score of 0.809 to 0.918. We can see that the performance for all noise ratios significantly improved compared to Table \ref{tab:noise_performance}. Only for $p_1=0.1$, there is no improvement.

We analyse the epoch-wise labels based on the probability $p_1=0.2$. According to Figure \ref{fig:label_attack}, the labels can be categorized in four groups. The probability of clean and corrupted labels for one epoch can be predicted with

\vspace{-3px}
\begin{equation}
 \;\;\;\; p_{clean}=p_{cl|cl}+p_{cl|co}=(1-p_1)*(1-p_2)+p_1*p_2=0.6+0.05=0.65,
\label{eq:p3}
\end{equation}
\begin{equation}
  \;  p_{corrupt}=p_{co|cl}+p_{co|co}=(1-p_1)*p_2+p_1*(1-p_2)=0.2+0.15=0.35.
\label{eq:p4}
\end{equation}

The labels with probability $p_{cl|co}$, meaning that a noisy label is flipped to correct in an epoch, are responsible for the performance gain. We hypothesize that the additional epoch-wise noise with probability $p_{co|cl}$ is widely robust during training.

\section{Discussion}

We observed that individual label attacks help to improve the performance. Flip probability $p_2$ was calculated according to the binomial distribution. The number of epochs for the prediction varied during training depending, e.g., on the label noise. As we could not find out the exact number of epochs, we used the average duration over the past training runs. However, according to Table \ref{tab:noise_performance2}, the performance is robust for a wide range of $p_2$. 

Moreover, we artificially inserted label noise prior to training. However, label noise may be biased, e.g., the probability of label noise on difficult examples is higher. This bias was not considered in our experiments. Effects on the performance may vary when label noise is directly derived from the dataset.

\section{Conclusion}

We showed that more label noise in the training decreases the performance on tuberculosis classification. We implemented a robust method to increase the performance when training with noisy labels. By flipping certain labels in each epoch, a fraction of noisy labels were converted to correct labels. These examples contributed to the training such that the performance significantly increased. Furthermore, the epoch-wise flips were widely robust during training, no significant performance drops existed when the epoch-wise label conversion strategy was integrated. This method can be extended and applied on multi-label problems.

\bibliographystyle{bvm2020}

\bibliography{2809}

\begin{thebibliography}{1}

\bibitem{radiology_errors}
Bruno MA, Walker EA, Abujudeh HH.
\newblock Understanding and Confronting Our Mistakes: The Epidemiology of Error
  in Radiology and Strategies for Error Reduction.
\newblock RadioGraphics. 2015;35(6):1668--1676.

\bibitem{Rajpurkar2018DeepLF}
Rajpurkar P, Irvin J, Ball RL, et~al.
\newblock Deep learning for chest radiograph diagnosis: A retrospective
  comparison of the CheXNeXt algorithm to practicing radiologists.
\newblock PLOS Medicine. 2018 11;15(11):1--17.

\bibitem{Zhang2018GeneralizedCE}
Zhang Z, Sabuncu M.
\newblock Generalized Cross Entropy Loss for Training Deep Neural Networks with
  Noisy Labels.
\newblock In: Bengio S, Wallach H, Larochelle H, et~al., editors. Advances in
  Neural Information Processing Systems 31; 2018.  p. 8778--8788.

\bibitem{datasets}
Jaeger S, Candemir S, Antani S, et~al.
\newblock Two public chest X-ray datasets for computer-aided screening of
  pulmonary diseases.
\newblock Quantitative imaging in medicine and surgery. 2014 12;4:475--7.

\bibitem{Huang2016DenselyCC}
Huang G, Liu Z, Weinberger KQ, et~al.
\newblock Densely Connected Convolutional Networks.
\newblock 2017 IEEE Conference on Computer Vision and Pattern Recognition
  (CVPR). 2016; p. 2261--2269.

\bibitem{adam}
Kingma D, Ba J.
\newblock Adam: A Method for Stochastic Optimization; 2014. Available from:
  \url{http://arxiv.org/abs/1412.6980}.

\end{thebibliography}
% Bitte setzen Sie hier Ihre Beitragsnummer ein und benennen Sie
% die BibTeX-Datei ebenfalls auf Ihre Beitragsnummer um.
%Kontrollzeiledef
\marginpar{\color{white}E\articlenumber} % Zeile nicht verändern!
\end{document}